%% file: main.tex
\documentclass{article}

    \PassOptionsToPackage{numbers, compress}{natbib}

\usepackage[final]{neurips_2019}

\usepackage[utf8]{inputenc} 
\usepackage[T1]{fontenc}    
\usepackage{hyperref}       
\usepackage{url}            
\usepackage{booktabs}       
\usepackage{amsfonts}       
\usepackage{nicefrac}       
\usepackage{microtype}      
\usepackage{tikz}
\usepackage{amsmath}
\usetikzlibrary{arrows,positioning} 
\tikzset{
    >=stealth',
    punkt/.style={
           rectangle,
           rounded corners,
           draw=black, very thick,
           text width=6.5em,
           minimum height=2em,
           text centered},
    pil/.style={
           ->,
           thick,
           shorten <=2pt,
           shorten >=2pt,}
}

\title{\Large Deep Physiological State Space Model for Clinical Forecasting}

\author{%
  Yuan Xue, Denny Zhou, Nan Du, Andrew Dai, Zhen Xu, Kun Zhang and Claire Cui \\
  Google\\
  \texttt{yuanxue,dennyzhou, dunan, adai, zhenxu, kunzhang, claire@google.com} \\
}

\begin{document}

\maketitle

\begin{abstract}
Clinical forecasting based on electronic medical records (EMR) can uncover the temporal correlations between patients' conditions and outcomes from sequences of longitudinal clinical measurements. In this work, we propose an intervention-augmented deep state space generative model to capture the interactions among clinical measurements and interventions by explicitly modeling the dynamics of patients' latent states. Based on this model, we are able to make a joint prediction of the trajectories of future observations and interventions. Empirical evaluations show that our proposed model compares favorably to several state-of-the-art methods on real EMR data.

\end{abstract}
\input{tikz}
\input{intro}

\input{problem}
\input{model}
\input{method}

\input{exp}
\input{conclude}

\bibliography{main}
\bibliographystyle{unsrtnat}

\end{document}

%% file: tikz.tex
%
%
%
%

\usetikzlibrary{shapes}
\usetikzlibrary{fit}
\usetikzlibrary{chains}
\usetikzlibrary{arrows}

\tikzstyle{latent} = [circle,fill=white,draw=black,inner sep=1pt,
minimum size=20pt, font=\fontsize{10}{10}\selectfont, node distance=1]
\tikzstyle{obs} = [latent,fill=gray!25]
\tikzstyle{const} = [rectangle, inner sep=0pt, node distance=1]
\tikzstyle{factor} = [rectangle, fill=black,minimum size=5pt, inner
sep=0pt, node distance=0.4]
\tikzstyle{det} = [latent, diamond]

\tikzstyle{plate} = [draw, rectangle, rounded corners, fit=#1]
\tikzstyle{wrap} = [inner sep=0pt, fit=#1]
\tikzstyle{gate} = [draw, rectangle, dashed, fit=#1]

\tikzstyle{caption} = [font=\footnotesize, node distance=0] %
\tikzstyle{plate caption} = [caption, node distance=0, inner sep=0pt,
below left=5pt and 0pt of #1.south east] %
\tikzstyle{factor caption} = [caption] %
\tikzstyle{every label} += [caption] %

\tikzset{>={triangle 45}}


\newcommand{\factoredge}[4][]{ %
  \foreach \f in {#3} { %
    \foreach \x in {#2} { %
      \path (\x) edge[-,#1] (\f) ; %
    } ;
    \foreach \y in {#4} { %
      \path (\f) edge[->,#1] (\y) ; %
    } ;
  } ;
}

\newcommand{\edge}[3][]{ %
  \foreach \x in {#2} { %
    \foreach \y in {#3} { %
      \path (\x) edge [->,#1] (\y) ;%
    } ;
  } ;
}

\newcommand{\factor}[5][]{ %
  \node[factor, label={[name=#2-caption]#3}, name=#2, #1,
  alias=#2-alias] {} ; %
  \factoredge {#4} {#2-alias} {#5} ; %
}

\newcommand{\plate}[4][]{ %
  \node[wrap=#3] (#2-wrap) {}; %
  \node[plate caption=#2-wrap] (#2-caption) {#4}; %
  \node[plate=(#2-wrap)(#2-caption), #1] (#2) {}; %
}

\newcommand{\gate}[4][]{ %
  \node[gate=#3, name=#2, #1, alias=#2-alias] {}; %
  \foreach \x in {#4} { %
    \draw [-*,thick] (\x) -- (#2-alias); %
  } ;%
}

\newcommand{\vgate}[6]{ %
  \node[wrap=#2] (#1-left) {}; %
  \node[wrap=#4] (#1-right) {}; %
  \node[gate=(#1-left)(#1-right)] (#1) {}; %
  \node[caption, below left=of #1.north ] (#1-left-caption)
  {#3}; %
  \node[caption, below right=of #1.north ] (#1-right-caption)
  {#5}; %
  \draw [-, dashed] (#1.north) -- (#1.south); %
  \foreach \x in {#6} { %
    \draw [-*,thick] (\x) -- (#1); %
  } ;%
}

\newcommand{\hgate}[6]{ %
  \node[wrap=#2] (#1-top) {}; %
  \node[wrap=#4] (#1-bottom) {}; %
  \node[gate=(#1-top)(#1-bottom)] (#1) {}; %
  \node[caption, above right=of #1.west ] (#1-top-caption)
  {#3}; %
  \node[caption, below right=of #1.west ] (#1-bottom-caption)
  {#5}; %
  \draw [-, dashed] (#1.west) -- (#1.east); %
  \foreach \x in {#6} { %
    \draw [-*,thick] (\x) -- (#1); %
  } ;%
}

%% file: intro.tex
\section{Introduction}
\vspace{-2mm}

The wide adoption of electronic medical records (EMR) has resulted in the collection of an enormous amount of patient measurements over time in the form of time-series data. These retrospective data contain valuable information that captures the intricate relationships between patient conditions and outcomes, and present a promising avenue for improving patient healthcare.

Recently, machine learning methods have been increasingly applied to EMR data to predict patient outcomes such as  mortality or diagnosis~\cite{rajkomar18scalable, bcb2017-sha, Che2018-vj, Choi2015-ak, iclr16-lipton, aaai2018-song, aim2013-liu, aaai2016-liu, jaim2017-wu}. Yet, the integration of the prediction results into clinicians' workflows still faces significant challenges as the alerts generated by these  machine learning algorithms provide few insights into why the predictions are made, and how to act on the predictions. In this paper, we present a deep state space generative model, augmented with intervention forecasting, which provides a principled way to capture the interactions among observations, interventions, hidden patient states and their uncertainty. Based on this model, we are able to provide simultaneous forecasting of biomarker trajectories and guides clinicians with intervention suggestions. The ability to jointly forecast multiple clinical variables provides clinicians with a full picture of a patient’s medical condition and better supports them with decision making.


%% file: problem.tex
\section{Learning Task and Model}
\label{sec:problem}
\vspace{-2mm}

Consider a longitudinal EMR system with $N$ patients. We discretize and calibrate patient $i$'s longitudinal records to a time window $[1, T_i]$, where time $1$ and $T_i$ represent the time when the patient first and last interacts with the system. When the context is clear, we simplify notation $T_i$ with $T$. We consider two types of time series data in EMR: 1) \textbf{observations} $\mathbf{x}$, a real-valued vector of $O$-dimension. Each dimension corresponds to one type of clinical measurement including vital signs and lab results (e.g., mean blood pressure, serum lactate). We use $\mathbf{x}_{1:T}$ to denote the sequence of measurements at discrete time points $t = 1, ..., T$; 2) \textbf{interventions} $\mathbf{u}$, a real-valued vector of $I$-dimension. Each dimension corresponds to one type of clinical intervention,
and its value indicates the presence and the level of intervention such as the dosage of medication being administrated or the settings of a mechanical ventilator. Similarly, $\mathbf{u}_{1:T}$ denotes the sequence of interventions at $t = 1, \ldots, T$. At prediction time $t^*$, given the sequence of observations and interventions $\mathbf{x}_{1:t^*}$, $\mathbf{u}_{1:t^*}$, we estimate the distribution of observations $\mathbf{x}_{t^*+\tau}$ and interventions $\mathbf{u}_{t^*+\tau}$ where $\tau \in [1, H]$. $H$ denotes forecasting horizon.




%% file: model.tex
\begin{figure}[h]
\centering
        \begin{tikzpicture}[scale=0.7, transform shape]
        \node[latent] (z1) {$z_1$};
        \node[obs, below= of z1] (u1) {$u_1$};
        \node[obs,above= of z1] (x1) {$x_1$};
        \node[latent,right=of z1] (z2) {$z_2$};
        \node[obs, below=of z2] (u2) {$u_2$};
        \node[obs, above=of z2] (x2) {$x_2$};
        \node[latent,right=of z2] (z3) {$z_3$};
        \node[obs, below=of z3] (u3) {$u_3$};
        \node[obs, above=of z3] (x3) {$x_3$};
        \node[latent,right=of z3] (z4) {$z_4$};
        \node[obs, above=of z4] (x4) {$x_4$};
        \node[obs, below= of z4] (u4) {$u_4$};
	\edge{z1}{z2};
	\edge{z2}{z3};
	\edge{z3}{z4};
	\edge{z1}{x1};
	\edge{z2}{x2};
	\edge{z3}{x3};
	\edge{z4}{x4};
	\edge{u1}{z1};	
	\edge{u2}{z2};
	\edge{u3}{z3};
	\edge{u4}{z4};
	\edge{z1}{u2};
	\edge{z2}{u3};
	\edge{z3}{u4};
    \end{tikzpicture}
    \caption{Graph Model For Patient Physiological State.}
\label{fig:gssm}
\end{figure}
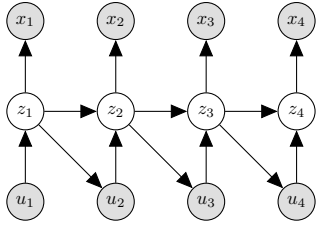

To provide a joint forecast, we need a powerful model that captures the temporal correlations among observations and interventions. To this end, we adopt a Gaussian state space model to explicitly model the latent patient physiological state as shown in Fig.~\ref{fig:gssm}. Let $\mathbf{z}_t$ be the latent variable vector that represents the physiological state at time $t$ and $\mathbf{z}_{1:T}$ be the sequence of such latent variables. The system dynamics are defined as:
\begin{align}
	&p(\mathbf{z}_t|\mathbf{z}_{t-1}, \mathbf{u}_{t}) \sim \mathcal{N} (\mathcal{A}_t(\mathbf{z}_{t-1}) + \mathcal{B}_t(\mathbf{u}_t), \bf{Q}) & \mathrm{Transition} \label{eqn:state_tran} \\ 
	&p(\mathbf{x}_t|\mathbf{z}_t) \sim \mathcal{N} (\mathcal{C}(\mathbf{z}_t), \bf{R}) &\mathrm{ Emission} \label{eqn:obs_emission}
\end{align}
where Eq.~\eqref{eqn:state_tran} defines the state transition: function $\mathcal{A}$ defines the system transition without external influence, i.e., how patient state will evolve from $\mathbf{z}_{t-1}$ to $\mathbf{z}_t$ without intervention. $\mathcal{B}$ captures the effect of intervention $\mathbf{u}_t$ on patient state $\mathbf{z}_t$. In Eq.~\eqref{eqn:obs_emission}, $\mathcal{C}$ captures the relationship between internal state $\mathbf{z}_t$ and observable measurements $\mathbf{x}_t$. $\mathbf{Q}$ and $\mathbf{R}$ are process and measurement noise covariance matrices. We assume them to be time-invariant. Eq.~\eqref{eqn:state_tran} and \eqref{eqn:obs_emission} subsume a large family of linear and non-linear state space models. For example, by setting $\mathcal{A}, \mathcal{B}, \mathcal{C}$ to be matrices, we obtain linear state space models. By parameterizing $\mathcal{A}, \mathcal{B}, \mathcal{C}$  via deep neural networks, we have deep state space models. 

\textit{Intervention Forecast}. Contrary to classical state space models, where interventions are usually considered as external factors, when inferring patient states from EMR data, interventions are an integral part of the system, as they are determined by clinicians based on their estimation of patient states and medical knowledge/clinical guidelines. To model this relationship, we augment the state space model with additional dependency from $\mathbf{z}_t$ to $\mathbf{u}_{t+1}$ as shown in Fig.~\ref{fig:gssm}.
\begin{equation}
p(\mathbf{u}_t|\mathbf{z}_{t-1}) \sim \mathcal{N}(\mathcal{D}(\mathbf{z}_{t-1}), \bf{U})  \label{eqn:int_emission}
\end{equation}
Similarly, in Eq.(\ref{eqn:int_emission}) $\mathcal{D}$ can be either a matrix for a linear model or parameterized by a neural network for a nonlinear model. $U$ is the intervention covariance. 

%% file: method.tex
\newcommand{\pth}{p_\theta}
\newcommand{\qph}{q_\phi}
\newcommand{\xfull}{\mathbf{x}_{1:T}}
\newcommand{\zfull}{\mathbf{z}_{1:T}}
\newcommand{\utwofull}{\mathbf{u}_{2:T}}
\newcommand{\ufull}{\mathbf{u}_{1:T}}
\newcommand{\x}{\mathbf{x}}
\newcommand{\z}{\mathbf{z}}
\newcommand{\sumt}{\sum_{t=1}^T}
\newcommand{\sumtm}{\sum_{t=1}^{T-1}}
\newcommand{\xh}{\mathbf{x}_{1:t^*}}
\newcommand{\zh}{\mathbf{z}_{1:t^*}}
\newcommand{\uh}{\mathbf{u}_{1:t^*}}
\newcommand{\xf}{\mathbf{x}_{t^*+1:t^*+\tau}}
\newcommand{\zf}{\mathbf{z}_{t^*+1:t^*+\tau}}
\newcommand{\uf}{\mathbf{u}_{t^*+1:t^*+\tau}}
\newcommand{\sumf}{\sum_{t=t^*+1}^{t^*+\tau}}
\newcommand{\Exp}[2]{\mathop{\mathbb{E}}_{#2}\left[#1\right]}

\section{Method}
\vspace{-2mm}

Our state space model is fully specified by the generative parameter $\theta = (\mathcal{A}, \mathcal{B}, \mathcal{C}, \mathcal{D})$. In this section, we present two learning learning objectives and their associated variational lower bounds that support the clinical forecast tasks as described in Sec.~\ref{sec:problem}. We also present the algorithm and the neural network models used for learning.

\subsection{System Identification}
\vspace{-2mm}

One classical method of estimating these parameters is to maximize the data likelihood in the entire patient record. We consider maximizing the joint likelihood of observations and interventions. Note that the objective here is slightly different from the learning of classical state space model which maximizes the conditional likelihood of observations given interventions~\cite{nips16-marco-srnn,nips17-marco-kvae,nipsw15-dkf}. This task is referred to as system identification.
\begin{equation}\label{eqn:system_id}
 \log \pth(\xfull, \utwofull) = \log \int_z \pth(\xfull, \utwofull, \zfull)
\end{equation}

This log likelihood is intractable when inferring the posterior $\pth(\zfull|\xfull, \utwofull)$. We adopt the variational inference method by introducing a variational distribution $\qph$ that approximates this posterior. To simply the notations, we assume $\mathbf{u}_1$ to be a fixed zero vector and use $\x$ for $\xfull$, $\mathbf{u}$ for $\ufull$, and $\z$ for $\zfull$. We optimize the evidence lower bound (ELBO) given as follows:
\begin{equation}\label{eqn:varlowbnd}
 \log \pth(\x, \mathbf{u})
  \geq \Exp{\log \pth(\x|\z)}{\qph(\z|\x,\mathbf{u})} + \Exp{\log \pth(\mathbf{u}|\z)}{\qph(\z|\x,\mathbf{u})} - \mathbb{KL}(\, \qph(\z|\x, \mathbf{u}) || \pth(\z|\x, \mathbf{u})\, )
\end{equation}
Similar to~\cite{nipsw15-dkf}, this ELBO can be factorized along time as:
\begin{equation}\label{eqn:facterized_varlowbnd}
\sumt \Exp{\log \pth(\x_t|\z_t)}{\qph(\z_t|\x,\mathbf{u})} + \sumtm\Exp{\log \pth(\mathbf{u}_{t+1}|\z_t)}{\qph(\z_t|\x,\mathbf{u})} - \sum_{t=2}^T \mathbb{KL}(\, \qph(\z_t| \z_{t-1}\x, \mathbf{u}) || \pth(\z_t| \z_{t-1}, \mathbf{u}_{t-1})\, )
\end{equation}

The lower bound in Eq.(\ref{eqn:facterized_varlowbnd}) has two components: 1) the reconstruction loss for both observations and interventions; 2) the regularization loss which measures the difference between the encoder and the simple prior distribution of the  latent state $\z$ given the transition model between $z_{t-1}$ and $z_t$ as defined in the state space model (Eq.(\ref{eqn:state_tran})).

\subsection{Trajectory Forecast}
\vspace{-2mm}

While the system identification task tries to capture the inherit dynamics of a patient, it does not directly optimize for forecasting the values of the observations at a given time point $t^*$ over the next $\tau$ period, unless the system dynamics are homogeneous. Here we present an explicit model for trajectory forecast by maximizing the joint likelihood of observation and intervention in the forecast horizon $[t^*+1, t^*+\tau]$, given their historical values within time range $[1, t^*]$. The joint likelihood, the corresponding ELBO and its time-factorized form are provided below. To simply the notations, we use $\vec{\x}, \vec{\mathbf{u}}$ to represent the forecast value $\vec{\x}_{t^*+1:t^*+\tau}, \vec{\mathbf{u}}_{t^*+1:t^*+\tau}$, $\bar{\x}, \bar{\mathbf{u}}$ to represent the historical value $\bar{\x}_{1:t^*}, \bar{\mathbf{u}}_{1:t^*}$, $\tilde{ \z}$ to represent $\z_{1:t^*+\tau}$ the latent state connecting history to the forecast horizon.
\begin{align}
\label{eqn:forecast_varlowbnd}
&\log \pth(\vec{\x}, \vec{\mathbf{u}} | \bar{\x}, \bar{\mathbf{u}}) = 
 \log \int_{\tilde{\z}} \pth(\vec{\x}, \vec{\mathbf{u}}, \tilde{\z} | \bar{\x}, \bar{\mathbf{u}}) = \log \int_{\tilde{\z}} \pth(\tilde{\z} | \bar{\x}, \bar{\mathbf{u}}) \pth(\vec{\x}|\tilde{\z}) \pth(\vec{\mathbf{u}}|\tilde{\z}) \\
 &\geq \Exp{\log \pth(\vec{\x}|\tilde{\z})}{\qph(\tilde{\z}| \bar{\x},\bar{\mathbf{u}})} + \Exp{\log \pth(\vec{\mathbf{u}}|\tilde{\z})}{\qph(\tilde{\z}| \bar{\x},\bar{\mathbf{u}})}  - \mathbb{KL}(\, \qph(\tilde{\z}| \bar{\x},\bar{\mathbf{u}}) || \pth(\tilde{\z} | \bar{\x}, \bar{\mathbf{u}}))\,\\
 &= \sumf \Exp{\log \pth(\x_t|\z_t)}{\qph(\z_t|\bar{\x}, \bar{\mathbf{u}})} + \sumf\Exp{\log \pth(\mathbf{u}_t|\z_{t-1})}{\qph(\z_t|\bar{\x}, \bar{\mathbf{u}})} \label{eq:forecast_loss}\\
 & - \sum_{t=1}^{t^* +\tau} \mathbb{KL}(\, \qph(\z_t| \z_{t-1}, \bar{\x}, \bar{\mathbf{u}}) || \pth(\z_t| \z_{t-1}, \bar{\mathbf{u}})\, )
\end{align}
The above forecast ELBO has two components: 1) the forecast loss for both observations and interventions over the forecast horizon (Eq.(\ref{eq:forecast_loss})); and 2) the regularization loss for latent state $\z$ from the history to the forecast horizon. Note that the encoder $\qph(\cdot)$ only depends on the historical values $\bar{\x}, \bar{\mathbf{u}}$ and rolls out the state for the future with their forecast values.

\subsection{Learning Algorithm and Model Architecture}
\vspace{-2mm}

Give the ELBOs of the above tasks, our learning algorithm proceeds the following steps: 1) inference of $\mathbf{z}$ from $\mathbf{x}$, and $\mathbf{{u}}$ by an encoder network $\qph$; 2) sampling based on the current estimate of the posterior $\mathbf{z}$ to either reconstruct the observation and the next step intervention (for system identification task), or forecast the next observation and the intervention afterwards (for trajectory prediction) based on the generative model $\pth$. For the latter case, the generative model will be used to roll out multiple time steps into the forecast horizon; 3) estimating gradients of the loss (negative ELBO) with respect to $\theta$ and $\phi$ and updating parameters of the model. Gradients are averaged across stochastically sampled mini-batches of the training set. We follow the same model architecture as in ~\cite{nipsw15-dkf} and use a LSTM as the encoder network, MLP for the state transition and observation emission. All models were implemented in TensorFlow~\cite{tensorflow2015-whitepaper} and the code will be open sourced.

%% file: exp.tex
\section{Experiments}
\vspace{-2mm}


We use Medical Information Mart for Intensive Care (MIMIC) data~\cite{johnson16mimic} in our empirical study. We select inpatients from MIMIC-III who are still alive $48$ hours after admission as our study cohort and forecast their vital signs and lab measurements jointly with interventions. There are $42026$ in-patient encounters included in the study with $3175$ observed in hospital death. We select the $96$ most frequently used observational data features and $8$ types of vasopressors and antibiotics, $6$ most recorded ventilation and dialysis machine settings as intervention features. All observation and intervention values are normalized using z-score.

Observational data is recorded at irregular intervals in EMR, resulting in a large number of missing values when sampled at regular time steps. We adopt a simple method where the most recent value is used to impute the missing ones for observations. For interventions, we need to differentiate the case where a missing value represents that the intervention is not performed or completed vs. the case where a missing value means the same setting is continued at this time step. Specifically, we pick the $90$-percentile at the distribution of inter-medication-administration time and the inter-intervention-setting time as the cut-off threshold. If two consecutive interventions are within the time range of their corresponding thresholds, then we consider the missing value as an indication of a continuous action and use the last setting for its missing value. If it falls outside of this range, then a missing value is considered as no action. 

The hyperparameters including the learning rate, the hidden state size for LSTM, the number of units and layers for MLP, the noise co-variance are tuned. The experiment uses a hidden state size of $50$ for LSTM and $32$ hidden units with $3$ layers for MLPs.

We use the mean absolute error (MAE) to evaluate the performance of trajectory prediction over different forecast horizons. We use $10$-fold cross validation and estimate the standard error of the mean.
For each fold, we split the dataset into train/eval/test according to $80\%$/$10\%$/$10\%$ based on the hash value of the patient ID. We compare the following models in our study:

\begin{itemize}
\item \textbf{History rollout (HR)} is a baseline model, which follows the method in ~\cite{nips18-amazon}. It trains a deep state space model based on the historical observations before the prediction time and rolls out the state predictions in the forecast horizon. 
\item \textbf{Kalman Filter (KF)}~\cite{kf} provides a baseline of linear forecast model. In this method, the generative parameters $\theta = (\mathcal{A}, \mathcal{B}, \mathcal{C}, \mathcal{D})$ are all matrices. The posterior state estimation of $\mathbf{z}$ is performed via close-form formula. 
\item \textbf{Trajectory forecast (TF)} is another baseline which directly uses the trajectory forecast ELBO defined in Eq.(\ref{eqn:forecast_varlowbnd}) to train the model.
\item \textbf{System identification + Trajectory forecast(SI+TF)} is our proposed method. Here we pretrain the deep state space model based on the system identification ELBO as defined in Eq.(\ref{eqn:facterized_varlowbnd}) then we fine tune the model based on the trajectory forecast loss (Eq.(\ref{eqn:forecast_varlowbnd})). 
\end{itemize}

\begin{table}[h!]
\centering
\vspace{-5mm}
\begin{tabular}{lp{2cm} p{2cm} p{2cm} p{2cm}}  \\
\toprule 
                & MAE@24hr & MAE@48hr &  MAE@72hr  \\
 \midrule                 
History rollout (HR) &	0.473(0.019)&	0.492(0.021)& 0.571(0.037)\\ 
Kalman Filter (KF)&	0.614(0.036) &	0.622(0.045)&	0.731(0.053) \\ 
Trajectory forecast (TF)  &	0.512(0.017)&0.528(0.019)&	0.546(0.022) \\ 
System identification (SI+TF) &	\textbf{0.453(0.012)} &	\textbf{0.453(0.012)}&		\textbf{0.514(0.020)}\\ 
 \bottomrule
\end{tabular}
\caption{\small Trajectory Forecast Results. Parentheses denote standard error.}
\label{tab:forecast}
\end{table}

The results in Table~\ref{tab:forecast} show that \textbf{SI+TF} consistently outperforms all baselines over all forecast horizons. For all methods, the forecast error gracefully increases with the length of forecasting horizon. As a linear baseline \textbf{KF} performs the worst, which demonstrates the predictive power of deep state space model. As forecast horizon increases from 24hr to 72hr, \textbf{HR} shows the largest performance penalty ($0.09$) among all the deep models, while \textbf{TF} has a penalty around $0.03$, as \textbf{TF} optimizes the future measurement likelihood directly, but \textbf{HR} relies on the consistency of the dynamics from the history to the future. 

%% file: conclude.tex
\section{Conclusion}
\vspace{-2mm}

In this work, we present a joint prediction of clinical measurement and intervention trajectories with the progression of the patient condition. Our prediction model is built upon on the deep state space model of patient physiological state, which provides a principled way to capture the interactions among observations, interventions and physiological state. Experiment study over MIMIC datasets shows that our proposed outperforms the state-of-art methods.